\pdfoutput=1
\documentclass[letterpaper]{article}
\usepackage[preprint]{aaai2027}
\usepackage[hyphens]{url}
\usepackage{graphicx}
\usepackage{natbib}
\usepackage{caption}
\usepackage{amsmath}
\usepackage{amssymb}
\usepackage{booktabs}
\usepackage{xspace}

\newcommand{\bench}{H$^2$U3D\xspace}

\usepackage{makecell}
\usepackage{pifont}
\usepackage{tcolorbox}
\tcbuselibrary{breakable,listings}

\newcommand{\cmark}{\ding{51}}
\newcommand{\xmark}{\ding{55}}

\newtcolorbox{codebox}[1][]{
  colback=gray!10,
  colframe=black,
  boxrule=0.4pt,
  arc=2pt,
  top=2pt,
  bottom=2pt,
  left=3pt,
  right=3pt,
  boxsep=3pt,
  fontupper=\small,
  listing only,
  breakable,
  listing options={
    basicstyle=\ttfamily\footnotesize,
    language=Python,
    keywordstyle=\color{blue},
    stringstyle=\color{red},
    commentstyle=\color{gray},
    morekeywords={\quad}
  },
  #1
}

\title{Think, Then Look: Active Spatial Reasoning for House-Scale 3D Scene Understanding}
\author{Hongpei Zheng\textsuperscript{\rm 1}, Shijie Li\textsuperscript{\rm 2}, Lin Qian\textsuperscript{\rm 1}, Zhenghao Li\textsuperscript{\rm 1}, Qijun Yang\textsuperscript{\rm 1}, Yanran Li\textsuperscript{\rm 3}, Hujun Yin\textsuperscript{\rm 1}}
\affiliations{\textsuperscript{\rm 1}University of Manchester\\\textsuperscript{\rm 2}I2R, A*STAR\\\textsuperscript{\rm 3}University of Bedfordshire}

\begin{document}

\maketitle

\begin{abstract}
Spatial reasoning in large-scale 3D environments remains challenging for
current vision--language models, which are typically constrained to room-scale
scenarios. We formalize \textbf{Active House-Scale Spatial Reasoning (AHSR)},
a new paradigm in which a model reasons over a pre-built house-scale 3D map
via virtual spatial tool invocations to answer spatial questions, without
exhaustive scene-wide processing. To support AHSR research, we introduce
\bench (Holistic House Understanding in 3D), the first benchmark targeting
house-scale 3D scene understanding, featuring environments with an average
aggregate floor area of 250.8\,m$^2$ and up to three floors, together with
hierarchical coarse-to-fine visual representations.
Building on \bench, we propose SpatialReasoner, an AHSR framework trained via
supervised fine-tuning with self-correction, followed by reinforcement learning
with a task-aware adaptive exploration reward. SpatialReasoner achieves state-of-the-art performance on
\bench with 64.9\% overall accuracy, outperforming strong baselines including
GPT-5.4 and Gemini-3.5-Flash, and generalizes effectively to MT-HM3D and
HM-EQA. These results demonstrate the clear advantage of active map-directed
exploration over passive scene-wide processing in house-scale 3D understanding.
\end{abstract}
\section{Introduction}
\label{sec:intro}

\begin{figure*}[t]
    \centering
    \includegraphics[width=0.85\textwidth,trim=4cm 2cm 4cm 3.8cm,clip]{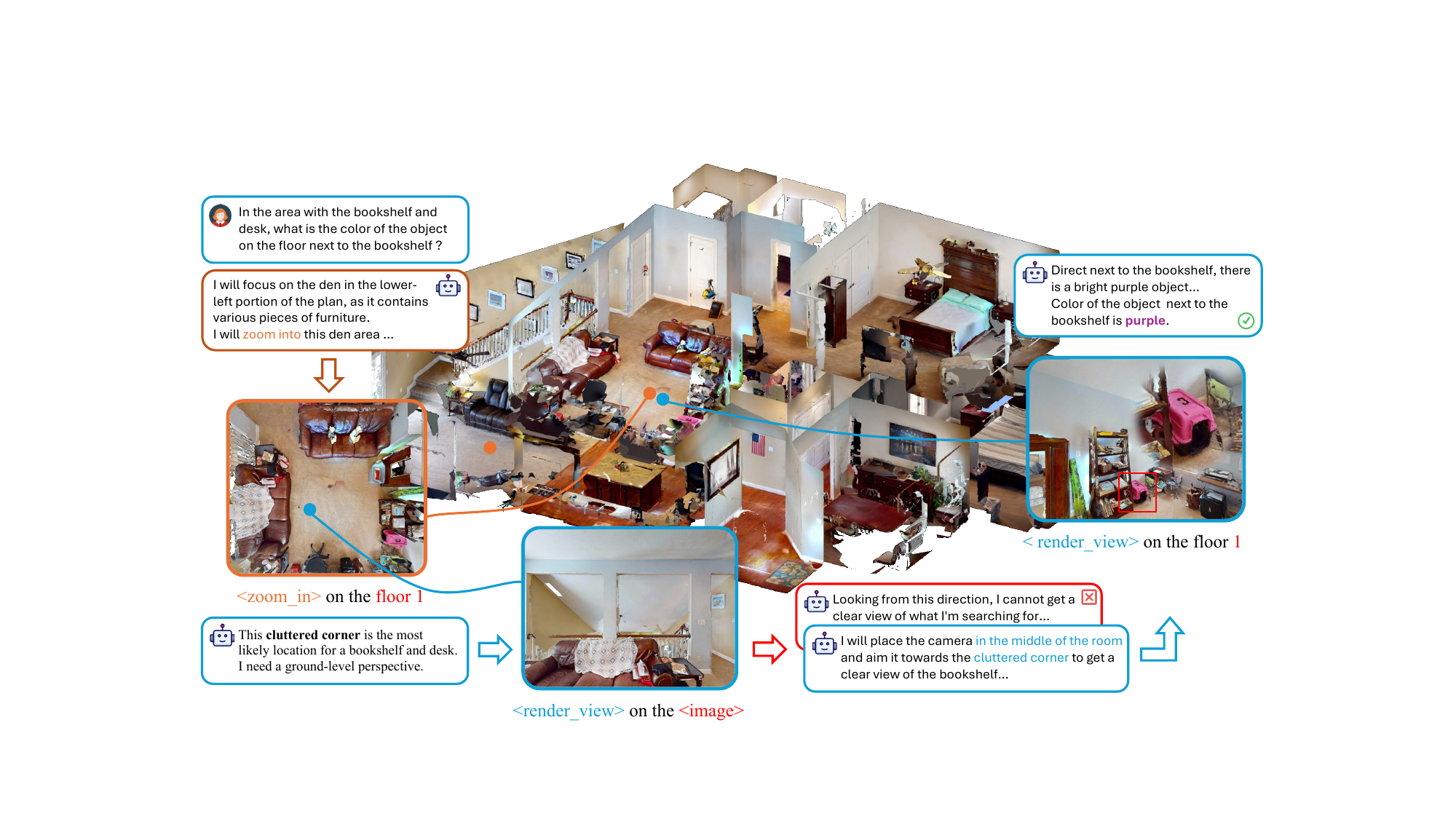}
    \caption{\textbf{Overview of SpatialReasoner inference.} Given a spatial question and a pre-built house-scale 3D map, the model progressively localizes the query-relevant region, invokes zooming and viewpoint-rendering tools, recovers from an uninformative observation, and produces an answer grounded in the final visual evidence.}
    \label{fig:inference_overview}
\end{figure*}

In embodied AI, agents operating over extended periods continuously build and
maintain an internal 3D representation of their surroundings. Such
representations are inherently house-scale---spanning multiple floors and dozens
of rooms---and serve as a persistent spatial memory for downstream reasoning
tasks. Efficiently answering natural language queries grounded in this
house-scale 3D memory remains an open and practically important problem.

Many existing 3D QA methods follow a \textbf{passive perception}
paradigm, processing a fixed, query-independent set of sampled views
or a scene-wide representation. Although effective in bounded scenes,
maintaining fine-grained coverage becomes increasingly costly as the
environment grows, requiring more observations and introducing more
query-irrelevant content. \textbf{Embodied QA} instead acquires
evidence through sequential navigation, typically without access to a
pre-built persistent map. Its observations are therefore constrained
by the agent's current pose and the navigable transitions required to
reach a target region.

A key observation motivates our approach: \textit{regardless of scene scale,
the region relevant to a given query remains spatially localized}. This suggests
that exhaustive scene-wide processing is unnecessary, and that selectively
directing perception toward query-relevant regions is both sufficient and
efficient. We formalize this insight as \textbf{Active House-Scale Spatial
Reasoning (AHSR)}: a paradigm in which a model, equipped with spatial tools
and a pre-built house-scale 3D map, actively localizes and inspects
query-relevant regions to answer spatial questions. The house-scale setting is central to AHSR---it is precisely because
agents operate persistently in such environments that a maintained map exists,
and it is at this scale that passive perception becomes computationally
intractable. AHSR is thus orthogonal to passive perception, which lacks
selectivity, and to Embodied QA, which operates without a prior map. Furthermore, while EQA is typically tethered to sequential physical exploration, AHSR enables random access to the environment via virtual spatial tools. By decoupling reasoning from the constraints of robotic movement, AHSR replaces slow navigation with efficient, query-driven evidence gathering.

To support research on AHSR, we introduce \bench (Holistic House
Understanding in 3D), a benchmark comprising 34,651 QA pairs across
797 HM3D scenes. The scenes have an average aggregate floor area of
250.8\,m$^2$ and contain up to three floors. Beyond question--answer
pairs, \bench provides hierarchical coarse-to-fine visual
representations and recorded spatial-operation annotations. A manually
verified subset of the training data further includes chain-of-thought
(CoT) annotations. These resources
support research on spatial question answering, exploration
decision-making, and tool-use reasoning.

Building on \bench, we propose SpatialReasoner, an AHSR framework that
autonomously invokes spatial tools---zooming into a region of interest or
rendering a first-person view from a specified viewpoint---to progressively
localize and inspect query-relevant areas within house-scale 3D environments.
We adopt a two-stage training strategy: supervised fine-tuning on
tool-interleaved trajectories augmented with self-correction examples to
instill the required output format and spatial operation patterns, followed
by reinforcement learning with spatially grounded rewards that adapt
exploration to rollout success, favor query-relevant spatial operations,
and penalize redundant tool invocations.

\begin{itemize}
    \item We identify and formalize \textbf{Active House-Scale Spatial Reasoning
    (AHSR)}, a new paradigm in which a model reasons over a pre-built house-scale
    3D map via virtual spatial tool invocations to answer spatial questions,
    without exhaustive scene-wide processing. AHSR fills the gap between passive
    perception, which lacks selectivity, and Embodied QA, which operates without
    a prior map.

    \item We introduce \bench, the first benchmark targeting house-scale 3D scene
    understanding, featuring scenes with an average aggregate
    floor area of 250.8\,m$^2$ and up to three floors.

    \item We propose SpatialReasoner, an AHSR framework trained via supervised
    fine-tuning with self-correction and reinforcement learning with a task-aware adaptive exploration
    reward. SpatialReasoner achieves the best overall accuracy of 64.9\% among
    the evaluated methods, outperforming strong baselines including GPT-5.4 and Gemini-3.5-Flash.
\end{itemize}
\section{Related Work}
\label{sec:related_work}

\noindent\textbf{Benchmarks for 3D Question Answering.}
Room-scale reconstructions such as ScanNet
~\cite{daiScanNetRichlyAnnotated3D2017}, ScanNet++
~\cite{yeshwanth2023scannethighfidelitydataset3d}, and
ARKitScenes~\cite{baruch2022arkitscenesdiverserealworlddataset}
have supported 3D question-answering benchmarks including ScanQA
~\cite{azumaScanQA3DQuestion2022} and SQA3D
~\cite{maSQA3DSituatedQuestion2023}. More recent benchmarks built on
HM3D~\cite{ramakrishnan2021hm3d}, including OpenEQA
~\cite{majumdar2023openeqa}, HM-EQA
~\cite{ren2024exploreconfidentefficientexploration}, and MT-HM3D
~\cite{zhai2025memorycentricembodiedquestionanswer}, extend question
answering to house-scale residential environments. However, their
evaluation episodes are restricted to a single navigable floor and do
not require reasoning over a jointly presented multi-floor map.

In contrast, \bench contains house-scale scenes with up to three floors
and jointly exposes the floor-level representations of each scene.
The model must first identify the query-relevant floor and region before
acquiring detailed visual evidence. To the best of our knowledge, \bench
is the first 3D question-answering benchmark designed for active
reasoning in multi-floor environments, with tool-interleaved annotations
that record floor selection, region zooming, and viewpoint rendering.

\noindent\textbf{Passive 3D Scene Reasoning.}
Most VLM-based approaches to 3D scene understanding operate on a
query-independent scene representation constructed before the question
is known. Multi-view methods such as Video3D-LLM
~\cite{zhengVideo3DLLMLearning2025} and 3D-LLaVA
~\cite{deng3DLLaVAGeneralist3D2025} reason over pre-sampled visual
observations, while related approaches incorporate 3D coordinate
information~\cite{zhuLLaVA3DSimpleEffective2025} or bird's-eye-view
representations~\cite{qiGPT4SceneUnderstand3D2025} to improve spatial
grounding. Although effective in bounded environments, these methods
process essentially the same visual evidence for every query. At house
scale, maintaining fine-grained scene coverage generally requires more
observations and introduces increasing amounts of query-irrelevant
information. Given a pre-built 3D map, we instead study query-time
evidence acquisition: the model uses coarse global context to identify
which regions and viewpoints require fine-grained inspection, avoiding
dense processing of detailed observations across the entire scene.

\noindent\textbf{Active Exploration and Spatial Memory.}
Active perception methods acquire query-relevant evidence through
interaction rather than relying solely on fixed visual inputs.
ExploreEQA~\cite{ren2024exploreconfidentefficientexploration} uses
answer confidence to guide exploration, while GraphEQA
~\cite{saxena2025grapheqausing3dsemantic} organizes observations with
3D semantic scene graphs. MemoryEQA
~\cite{zhai2025memorycentricembodiedquestionanswer} and 3D-Mem
~\cite{Yang_2025_CVPR} further maintain structured spatial memories to
support long-horizon reasoning. These methods primarily formulate
evidence acquisition as sequential embodied exploration, in which the
available observations depend on the agent's current pose and navigable
transitions. Struct2D
~\cite{zhu2025struct2dperceptionguidedframeworkspatial} also employs BEV
representations and reasoning annotations for spatial question answering.

AHSR considers a complementary setting in which the environment has
already been reconstructed into a persistent multi-floor 3D map.
The model can therefore access query-relevant regions through virtual
zooming and viewpoint-rendering operations. SpatialReasoner focuses on learning a
query-conditioned map-access policy, separating the decision of where
to inspect from the navigation required to reach that location.

\section{\bench Dataset}
\label{sec:problem_formulation}

In this section, we present \bench, Holistic House Understanding in 3D, a Visual Question Answering (VQA) dataset targeting house-scale 3D scene understanding (3DSU). \bench is built upon the HM3D \cite{ramakrishnan2021hm3d} dataset, which provides high-quality 3D reconstructions of large-scale residential environments. Compared to previous 3D VQA datasets \cite{azumaScanQA3DQuestion2022, maSQA3DSituatedQuestion2023}, which are limited to room-scale, \bench extends the scale to house-level, including multi-floor and multi-room environments, making the task significantly more challenging due to the much larger spatial scale. Specifically, each data sample in \bench is represented as $(\mathcal{S}, q, a)$, where $\mathcal{S}$ denotes a multi-floor, multi-room, ultra-large indoor scene, and $q$ is a natural language query related to $\mathcal{S}$; $a$ is the correct answer expected to be produced by the model.

\subsection{Task Formulation}

We define \textbf{Active House-Scale Spatial Reasoning (AHSR)} as
follows. Given a pre-built house-scale 3D map $\mathcal{S}$, its
floor-level BEVs $F$, a natural-language query $q$, and a set of virtual
spatial tools $\mathcal{T}$, a model iteratively invokes tools
$u_{1:T}$ to acquire query-relevant visual observations $o_{1:T}$ and
produces an answer $\hat{a}$ within a tool-call budget $B$. The model
initially observes only the coarse multi-floor context $F$; detailed
regional and first-person observations are selected during inference
conditioned on the query and the preceding interaction history.

Unlike passive 3D QA, AHSR does not fix the detailed visual evidence
before the query is known. To support the learning of such
query-conditioned map-access policies, each \bench example additionally
provides the reference floor, zoom region, viewpoint configuration, and
a tool-interleaved exploration trajectory. These annotations provide
training supervision rather than defining a unique valid exploration
path.

\subsection{Dataset Curation \& Annotations}

\begin{figure*}[t]
    \centering
    \includegraphics[width=\textwidth,trim=2.2cm 6cm 2.2cm 6cm,clip]{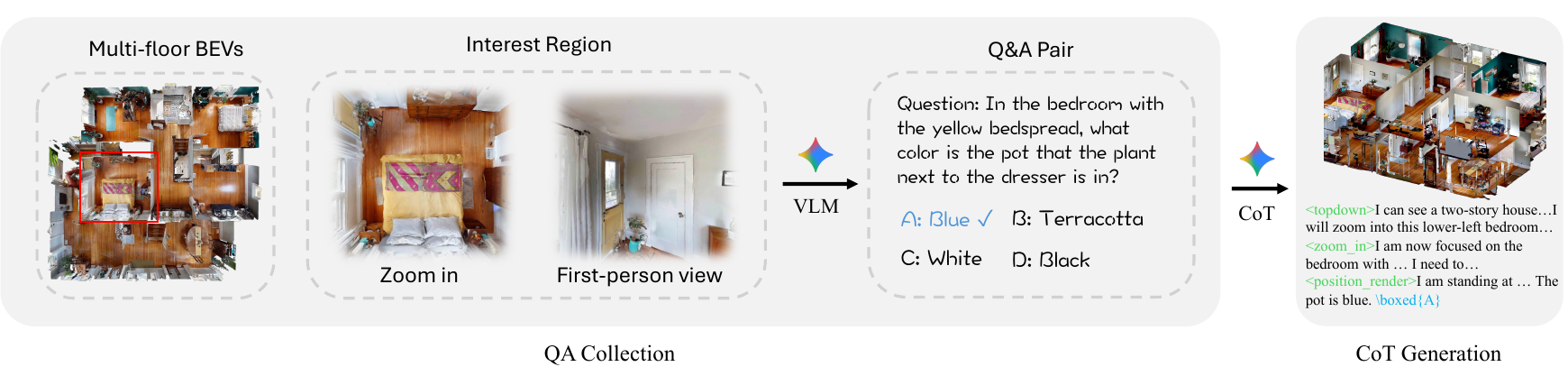}
    \caption{Two-stage \bench dataset construction pipeline. Stage 1 (QA Collection): tool calls are invoked to acquire interest region, then multi-floor BEVs and rendered views are processed by VLM to generate question-answer pairs. Stage 2 (CoT Generation): The same visual inputs and tool sequences are used to produce detailed chain-of-thought reasoning annotations.}
    \label{fig:dataset_construction}
\end{figure*}

\noindent \textbf{QA Data Generation.}
For each house-scale scene, we first use Habitat-Simulator to render a high-resolution BEV of every floor, providing the generation model with global context about the multi-floor layout.
Conditioned on these BEVs, Gemini-3.1-Pro \cite{googledeepmind2026gemini31pro} identifies a question-worthy region and invokes \texttt{zoom\_in} with a selected floor index and a two-dimensional bounding box.
After inspecting the returned local BEV, the model selects a ground-plane position and camera orientation within the region and invokes \texttt{render\_view} to obtain a first-person observation.
Views without sufficient visual information are discarded.

The model then generates a QA pair conditioned jointly on the multi-floor BEVs, the zoomed regional view, and the final first-person observation.
The generation prompt requires the question to be grounded in the explored scene, to identify its referents unambiguously, and to have exactly one answer supported by the visual evidence.
For every accepted QA pair, we retain the selected floor, zoom region, viewpoint position, and camera orientation, together with the complete tool history.

\noindent \textbf{Quality Control and Human Validation.}
To reduce generation errors and model-specific bias, we use GPT-5.2 \cite{singh2026openaigpt5card} to screen all QA candidates generated by Gemini-3.1-Pro for answerability, ambiguity, and answer correctness.
We then apply deduplication and category filtering and divide the data into scene-disjoint training, validation, and test sets.

Every QA pair in the validation and test sets is independently reviewed by two annotators using the corresponding scene evidence.
Each annotator judges whether the question is answerable, whether its interpretation and referents are unambiguous, and whether the annotated answer is correct.
A sample is retained only when both annotators consider it valid.

The annotators agreed on 82.37\% of the reviewed samples, and 1493 samples were removed during human validation.

\noindent \textbf{CoT Annotation.}
We augment a subset of the accepted training QA pairs with tool-interleaved reasoning annotations.
For each selected sample, we provide Gemini-3.1-Pro with the question and annotated answer, the multi-floor BEVs, the zoomed regional view, the rendered first-person observation, and the complete sequence of tool invocations and their parameters.
The model is instructed to produce a reasoning trace aligned with the recorded exploration trajectory.

The annotation describes the purpose of each spatial operation, the visual evidence obtained from its returned observation, and how that evidence informs the subsequent operation or final prediction.
The resulting trajectory interleaves reasoning steps with the corresponding \texttt{zoom\_in} and \texttt{render\_view} calls and terminates with an answer grounded in the collected visual evidence.
All generated CoT annotations are manually inspected before being included in the training data.

\subsection{Statistics \& Analysis}

After deduplication and category filtering, \bench contains 34,651 multiple-choice QA pairs collected from 797 HM3D scenes.
The dataset is divided at the scene level into training, validation, and test sets, containing 27,674, 3,605, and 3,372 QA pairs from 638, 80, and 79 scenes, respectively, with no scene overlap across splits.
Each QA record contains the scene and floor identifiers, four answer choices, the correct answer, and its question category, together with the associated zoom region, ground-plane viewpoint, and camera orientation.
Among the 27,674 training QA pairs, 5,626 samples covering 635 training scenes are further augmented with Chain-of-Thought (CoT) annotations.

The questions cover six categories: Position (9,838; 28.4\%), Attribute (9,416; 27.2\%), Counting (5,147; 14.9\%), Material (4,666; 13.5\%), State (2,846; 8.2\%), and Shape (2,738; 7.9\%).

\begin{figure}[htbp]
    \centering
    \includegraphics[width=\columnwidth]{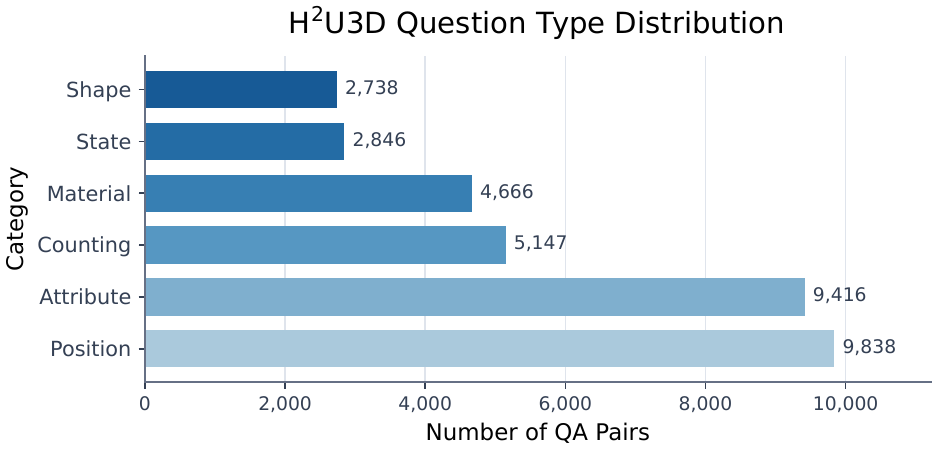}
    \caption{Question-category distribution of \bench.}
    \label{fig:h2u3d_category_distribution}
\end{figure}

The question-category distribution is shown in Fig.~\ref{fig:h2u3d_category_distribution}.
On average, each scene spans 250.8\,m$^2$ and contains 1.73 floors.
More details about the dataset construction process can be found in the the Appendix.

\section{SpatialReasoner}

As illustrated in Fig.~\ref{fig:inference_overview}, SpatialReasoner is an active reasoning framework for house-scale 3D scene understanding.
Given a pre-built 3D map and a textual query, the model starts from multi-floor BEVs and iteratively invokes region-zooming and viewpoint-rendering tools to acquire query-relevant visual evidence.
The returned observations are incorporated into the reasoning context until the model produces a grounded final answer.
We first formalize this tool-interleaved inference process and then describe its two-stage training procedure, consisting of supervised fine-tuning with self-correction and reinforcement learning with spatially grounded rewards.

\subsection{Formal Definition}
\label{sec:form}

Under the AHSR formulation, SpatialReasoner starts from the floor-level
BEVs $F$ of map $\mathcal{S}$ and accesses the map through two spatial
tools. $\texttt{zoom\_in}(f_\ell,\mathbf{b})$ returns a magnified BEV
of region $\mathbf{b}$ on floor $\ell$, whereas
$\texttt{render\_view}(\mathbf{p},\theta)$ returns a first-person
observation from ground-plane position $\mathbf{p}$ and camera yaw
$\theta$ on the currently selected floor.

At step $t$, the model samples

\begin{equation}
(z_t,u_t)
\sim
\pi_\phi\!\left(
\cdot\mid q,F,z_{<t},u_{<t},o_{<t}
\right),
\end{equation}

where $z_t$ is an intermediate reasoning segment and $u_t$ is either
a tool call in $\mathcal{T}$ or a final action
$\texttt{answer}(\hat{a})$. For a tool call, the environment returns
$o_t=\mathcal{E}(\mathcal{S},u_t)$ and appends it to the context; an
answer action terminates the interaction. The process permits at most
$B$ tool calls.

\subsection{Supervised Fine-Tuning with Self-Correction}
\label{sec:sft}

SpatialReasoner is initialized from Qwen3-VL-8B-Instruct
\cite{yang2025qwen3technicalreport}. Although the base model supports
generic tool use, it has not been trained for our spatial tool interface
or the associated coarse-to-fine interaction protocol. We therefore
perform supervised fine-tuning on the tool-interleaved
CoT subset of \bench to align the model with the required output format
and spatial operation patterns.

\noindent \textbf{Error-Induced Self-Correction Trajectories.}
Because the verified CoT trajectories primarily demonstrate successful
exploration, they provide limited supervision for recovering from
ineffective tool calls. We therefore augment 25\% of them with
template-based error-and-correction sequences. Starting from a verified
trajectory, we perturb either its zoom region, ground-plane position,
or camera orientation and execute the perturbed operation in the
rendering environment. We then insert the returned observation,
a failure-analysis step, and corrective tool calls that reconnect the
sequence to the original successful trajectory.

For 70\% of the augmented trajectories, progressive correction refines
the operation through two or three attempts. The remaining 30\% use
reset correction, abandoning the ineffective configuration and
restarting from an alternative region or viewpoint. Both strategies
eventually recover valid visual evidence and preserve the original
answer.

During SFT, the injected erroneous operation and its returned
observation are retained in the model context but excluded from the
prediction targets. We optimize the masked language-modeling objective

\begin{equation}
\mathcal{L}_{\mathrm{SFT}}(\phi)
=
-\sum_j m_j
\log P_\phi
\left(
y_j\mid y_{<j},q,V
\right),
\label{eq:sft_objective}
\end{equation}

where $V$ denotes the interleaved visual observations. We set $m_j=0$
for the injected erroneous operations and environment-returned outputs,
and $m_j=1$ for the failure analysis, corrective actions, subsequent
reasoning, and final answer. The model thus learns to recover from an
erroneous context without being trained to reproduce the injected
error.

\subsection{Reinforcement Learning with Spatial Rewards}
\label{sec:rl}

Supervised fine-tuning equips the model with the required output structure, spatial tool-use patterns, and self-correction behaviors.
We initialize reinforcement learning from this checkpoint and, for each input $(F,q)$, roll out a group of $G$ tool-interleaved reasoning trajectories $\{y_1,y_2,\ldots,y_G\}$.
During each rollout, every predicted spatial operation is immediately executed in Habitat-Simulator, and the returned visual observation is appended to the interaction context before the model makes its next decision.
Each completed trajectory contains intermediate reasoning, spatial operations, environment observations, and a final answer.
We score these trajectories using the composite reward defined below and optimize the policy with Group Relative Policy Optimization (GRPO) \cite{shao2024deepseekmathpushinglimitsmathematical} based on their relative rewards within each group.

\subsubsection{Adaptive Exploration Reward
($R_{\text{explore}}$)}

A fixed exploration bonus may encourage unnecessary tool calls,
whereas a fixed efficiency penalty may suppress information
acquisition before sufficient evidence has been collected. We
therefore adapt the exploration incentive to the rollout-group success
rate
$c(q)=G^{-1}\sum_{i=1}^{G}R_{\text{correct}}(y_i)$, where
$R_{\text{correct}}$ is binary. For trajectory $y_i$, let
$N_u(y_i)=N_{\text{uz}}(y_i)+N_{\text{ur}}(y_i)$ denote the number of
distinct zoomed regions and rendered viewpoints.

We define the interpolation coefficient and adaptive exploration
reward as

\begin{equation}
\begin{aligned}
\lambda(c)
&=
\min\left(
1,
\max\left(
0,
\frac{\tau_{\text{high}}-c}
{\tau_{\text{high}}-\tau_{\text{low}}}
\right)
\right), \\
R_{\text{explore}}(y_i;c)
&=
\lambda(c)\alpha_{\text{e}}
\min\left(N_u(y_i),N_{\max}\right) \\
&\quad-
\left(1-\lambda(c)\right)\alpha_{\text{p}} \\
&\qquad\times
\max\left(0,N_u(y_i)-N_{\text{budget}}\right).
\end{aligned}
\end{equation}

Here $\alpha_{\text{e}},\alpha_{\text{p}}>0$ control the exploration
bonus and efficiency penalty, $N_{\max}$ caps the bonus, and
$N_{\text{budget}}$ sets the efficiency threshold. Low group success
places more weight on evidence acquisition; as success increases, the
reward shifts toward penalizing exploration beyond the budget.

To discourage redundant evidence acquisition, let
$n_{\text{rep}}(y_i)$ count near-duplicate operations. A zoom is
repetitive when it selects the same floor and overlaps a previous
region above an IoU threshold; a rendered view is repetitive when its
position and circular-direction differences from a previous view fall
below predefined thresholds. We apply

\begin{equation}
P_{\text{rep}}(y_i)=
-\frac{\alpha_{\text{rep}}}{2}
n_{\text{rep}}(y_i)
\left(n_{\text{rep}}(y_i)+1\right),
\end{equation}

where $\alpha_{\text{rep}}>0$. The budget term regulates exploration
extent, while $P_{\text{rep}}$ specifically penalizes repeated
evidence acquisition.

\subsubsection{Goal-Directed Spatial Reward
($R_{\mathrm{goal}}$)}

The adaptive exploration reward regulates how extensively the model
explores, but does not assess whether the acquired observations are
relevant to the query. We therefore use the spatial annotations retained
during dataset construction as soft reference targets. For each question,
let
$g^*=(f^*,\mathbf{b}^*,\mathbf{p}^*,\theta^*)$
denote the reference floor, zoom region, camera position, and viewing
direction. For trajectory $y_i$, let
$(\hat f_i^{\mathrm{z}},\hat{\mathbf{b}}_i)$ and
$(\hat f_i^{\mathrm{v}},\hat{\mathbf{p}}_i,\hat{\theta}_i)$
denote its final valid zooming and viewpoint-rendering operations.

With region and position coordinates normalized to $[0,1]$, we define
$d_{\mathrm{b}}=\frac{1}{2}
\lVert\hat{\mathbf{b}}_i-\mathbf{b}^*\rVert_2$
and
$d_{\mathrm{p}}=\frac{1}{\sqrt{2}}
\lVert\hat{\mathbf{p}}_i-\mathbf{p}^*\rVert_2$.
Let
$d_{\mathbb{S}^1}(\hat{\theta}_i,\theta^*)\in[0,1]$
denote the shortest circular angular distance normalized by
$180^\circ$. The goal-directed reward is

\begin{equation}
\begin{aligned}
R_{\mathrm{goal}}(y_i)={}&
\omega_{\mathrm{z}}
\mathbf{1}[\hat f_i^{\mathrm{z}}=f^*]
\exp\left(
-\frac{d_{\mathrm{b}}^2}{2\sigma_{\mathrm{b}}^2}
\right) \\
&+
\omega_{\mathrm{v}}
\mathbf{1}[\hat f_i^{\mathrm{v}}=f^*]
\exp\left(
-\frac{d_{\mathrm{p}}^2}{2\sigma_{\mathrm{p}}^2}
-\frac{
d_{\mathbb{S}^1}(\hat{\theta}_i,\theta^*)^2
}{2\sigma_{\theta}^2}
\right),
\end{aligned}
\label{eq:goal_reward}
\end{equation}

where $\omega_{\mathrm{z}}+\omega_{\mathrm{v}}=1$.
If an operation is missing or invalid, its corresponding term is set
to zero. Scoring the final valid operations allows the model to recover
from intermediate spatial errors, while redundant calls are handled
separately by $P_{\mathrm{rep}}$. The Gaussian kernels provide partial
credit to spatially close alternatives without requiring the policy to
exactly reproduce the reference trajectory.

\subsection{Overall Reward and Optimization}

For each trajectory $y_i$ sampled for question $q$, we combine the task outcome and spatial-behavior rewards as

\begin{equation}
\begin{aligned}
R(y_i;q) ={}&
w_{\text{c}}R_{\text{correct}}(y_i)
+w_{\text{e}}R_{\text{explore}}(y_i;c) \\
&+w_{\text{g}}R_{\text{goal}}(y_i)
+w_{\text{r}}P_{\text{rep}}(y_i),
\end{aligned}
\label{eq:overall_reward}
\end{equation}

where all reward weights are non-negative.
The correctness term evaluates the final task outcome, while the remaining terms regulate exploration extent, spatial relevance, and repetitive tool use, respectively.

For each question, we standardize the rewards of its $G$ sampled trajectories using the group mean and standard deviation to obtain group-relative advantages.
We then optimize the policy using the standard clipped GRPO objective with reference-policy KL regularization.
The graded spatial rewards distinguish trajectories that receive the same answer-correctness reward but differ in where and how efficiently they explore.

More implementation details are provided in the
the Appendix.
\section{Experiments}
\label{sec:exp}

\subsection{Experimental Setup}

\noindent\textbf{Datasets and Metrics.}
We conduct training and evaluation on \bench, which is divided at the scene level into 27,674 training questions from 638 scenes, 3,605 validation questions from 80 scenes, and 3,372 test questions from 79 scenes.
The validation set is used for model selection, while all main results are reported on the scene-disjoint test set.
We report multiple-choice accuracy for six question categories: Position, Attribute, Counting, Material, State, and Shape.
Overall denotes the accuracy computed over all test questions.
We additionally evaluate cross-benchmark generalization on MT-HM3D and HM-EQA, as well as two real-world environments reconstructed using 3D Gaussian Splatting.

\noindent\textbf{Implementation Details.}
SpatialReasoner is initialized from Qwen3-VL-8B-Instruct \cite{yang2025qwen3technicalreport}.
During supervised fine-tuning, we use 5,626 manually verified, tool-interleaved CoT examples from the training split, of which 25\% are augmented with error-induced self-correction trajectories.
We use AdamW \cite{loshchilovDecoupledWeightDecay2019} and train for 3 epochs with a global batch size of 16, a cosine learning-rate schedule, and a peak learning rate of $10^{-6}$.
For reinforcement learning, we use the remaining 22,048 training questions that are not included in the SFT subset.
For each question, we sample 8 trajectories with a temperature of 1.0 and allow at most 8 tool calls per trajectory.
GRPO training is performed for 1 epoch with a learning rate of $10^{-6}$.

\noindent\textbf{Baselines.}
We compare SpatialReasoner with three groups of baselines.
Tool-free MLLMs include Gemini-3.5-Flash \cite{googledeepmind2026gemini35flash}, GPT-5.4 \cite{singh2026openaigpt5card}, Qwen3-VL-8B, Qwen3-VL-30B-A3B, and Qwen3-VL-235B \cite{yang2025qwen3technicalreport}.

These models receive the question, answer choices, and the BEV image of each floor.
Passive 3D models include Video-3D-LLM
\cite{zhengVideo3DLLMLearning2025}, LLaVA-3D
\cite{zhuLLaVA3DSimpleEffective2025}, VG-LLM
\cite{zhengLearningVideos3D2025}, and Ross3D
\cite{wangRoss3DReconstructiveVisual2025}. Following their respective
input protocols, Video-3D-LLM, VG-LLM, and Ross3D receive 32 sampled
frames, while LLaVA-3D receives 24.
Active perception baselines include ExploreEQA (Prismatic-7B) \cite{ren2024exploreconfidentefficientexploration} and MemoryEQA (Qwen3-VL-8B) \cite{zhai2025memorycentricembodiedquestionanswer}.
We further equip Qwen3-VL-8B, Qwen3-VL-235B, and GPT-5.4 with our rendering interface as direct tool-use baselines.
For a controlled comparison, these models receive the same initial multi-floor BEVs, access the same region-zooming and viewpoint-rendering operations, and share a maximum budget of 8 tool calls per question.

\subsection{Experimental Results}

\begin{table*}[t]
\centering
\caption{Accuracy on the \bench test set. The best and second-best
results are highlighted in bold and underlined, respectively.}
\label{tab:main-results}
\footnotesize
\setlength{\tabcolsep}{3.8pt}
\renewcommand{\arraystretch}{1.00}
\begin{tabular*}{\textwidth}{@{\extracolsep{\fill}}clcccccccc@{}}
\toprule
\textbf{\#}
& \textbf{Model}
& \textbf{Params.}
& \textbf{Position}
& \textbf{Attribute}
& \textbf{Counting}
& \textbf{Material}
& \textbf{State}
& \textbf{Shape}
& \textbf{Overall} \\
\midrule
\multicolumn{10}{@{}l}{\textit{Tool-Free Models}} \\
1 & Gemini-3.5-Flash
& --- & \underline{0.604} & 0.483 & 0.510 & 0.608 & 0.557 & 0.604 & 0.553 \\
2 & GPT-5.4
& --- & 0.588 & 0.540 & \underline{0.566} & 0.584 & 0.543
& \textbf{0.648} & 0.571 \\
3 & Qwen3-VL-8B
& 8B & 0.474 & 0.424 & 0.336 & 0.502 & 0.550 & 0.576 & 0.458 \\
4 & Qwen3-VL-30B-A3B
& 30B & 0.398 & 0.329 & 0.391 & 0.452 & 0.550 & 0.580 & 0.412 \\
5 & Qwen3-VL-235B
& 235B & 0.462 & 0.453 & 0.377 & 0.478 & 0.595 & 0.596 & 0.471 \\
\midrule
\multicolumn{10}{@{}l}{\textit{Passive 3D Models}} \\
6 & Video-3D-LLM
& 7B & 0.360 & 0.392 & 0.242 & 0.447 & 0.392 & 0.484 & 0.375 \\
7 & LLaVA-3D
& 7B & 0.354 & 0.359 & 0.314 & 0.441 & 0.354 & 0.348 & 0.361 \\
8 & VG-LLM
& 8B & 0.517 & 0.519 & 0.488 & 0.621 & 0.440 & 0.600 & 0.527 \\
9 & Ross3D
& 7B & 0.571 & 0.534 & 0.492 & \textbf{0.707} & 0.526 & 0.564 & 0.563 \\
\midrule
\multicolumn{10}{@{}l}{\textit{Active and Tool-Use Models}} \\
10 & ExploreEQA (Prismatic-7B)
& 7B & 0.487 & 0.462 & 0.385 & 0.619 & 0.498 & 0.364 & 0.475 \\
11 & MemoryEQA (Qwen3-VL-8B)
& 8B & 0.439 & 0.488 & 0.324 & 0.570 & 0.426 & 0.636 & 0.467 \\
12 & Qwen3-VL-8B w/ Tool
& 8B & 0.481 & 0.482 & 0.338 & 0.535 & 0.485 & 0.608 & 0.478 \\
13 & Qwen3-VL-235B w/ Tool
& 235B & 0.563 & 0.534 & 0.496 & 0.599 & \underline{0.612} & 0.636 & 0.560 \\
14 & GPT-5.4 w/ Tool
& --- & 0.595 & \underline{0.593} & 0.559
& 0.652 & 0.605 & \underline{0.644} & \underline{0.601} \\
\cmidrule(lr){1-10}
15 & \textbf{SpatialReasoner (Ours)}
& 8B & \textbf{0.644} & \textbf{0.676} & \textbf{0.617}
& \underline{0.654} & \textbf{0.622} & \underline{0.644}
& \textbf{0.649} \\
\bottomrule
\end{tabular*}
\end{table*}

\begin{table}[t]
\centering
\caption{Accuracy on MT-HM3D and HM-EQA benchmarks.}
\footnotesize
\setlength{\tabcolsep}{4.5pt}
\renewcommand{\arraystretch}{1.05}
\begin{tabular*}{\columnwidth}{@{\extracolsep{\fill}}lccc@{}}
\toprule
\textbf{Model}
& \textbf{Base VLM}
& \textbf{MT-HM3D}
& \textbf{HM-EQA} \\
\midrule
ExploreEQA & GPT-4o & 0.362 & 0.474 \\
GraphEQA   & GPT-4o & --- & 0.635 \\
MemoryEQA  & GPT-4o & 0.431 & 0.614 \\
\midrule
\textbf{SpatialReasoner}
& Qwen3-VL-8B & \textbf{0.479} & \textbf{0.668} \\
\bottomrule
\end{tabular*}
\label{tab:other_results}
\end{table}

We report the main results on the \bench test set in
Table~\ref{tab:main-results}. SpatialReasoner achieves the highest
overall accuracy of 64.9\%, outperforming the strongest tool-enabled
baseline, GPT-5.4 w/ Tool, by 4.8 percentage points (64.9\% vs.
60.1\%). It also exceeds the strongest tool-free model, GPT-5.4, by
7.8 percentage points.

The strongest passive 3D baseline, Ross3D, trails SpatialReasoner by
8.6 percentage points overall (56.3\% vs. 64.9\%), further
demonstrating the advantage of query-directed exploration over fixed
sampled views in house-scale environments.

Direct tool access improves all three evaluated VLMs. Qwen3-VL-8B,
Qwen3-VL-235B, and GPT-5.4 gain 2.0, 8.9, and 3.0 percentage points,
respectively, compared with their tool-free counterparts. However,
tool access alone does not ensure effective spatial exploration.
More importantly, under the same Qwen3-VL-8B backbone and tool
interface, SpatialReasoner improves over the directly prompted
Qwen3-VL-8B w/ Tool baseline by 17.1 percentage points (64.9\% vs.
47.8\%). This controlled comparison demonstrates the importance of
supervised tool-use alignment and reinforcement post-training for
learning effective exploration strategies.

We further evaluate SpatialReasoner on MT-HM3D
\cite{zhai2025memorycentricembodiedquestionanswer} and HM-EQA
\cite{ren2024exploreconfidentefficientexploration}, as shown in
Table~\ref{tab:other_results}. SpatialReasoner achieves accuracies of
47.9\% and 66.8\%, outperforming the strongest compared baselines by
4.8 percentage points on MT-HM3D and 3.3 percentage points on HM-EQA,
respectively. These results demonstrate consistent improvements across
other established HM3D-based question-answering settings.

\subsection{Real-World Reconstruction Evaluation}

To evaluate SpatialReasoner beyond existing reconstructed-scene benchmarks, we conduct an additional evaluation on two real-world indoor environments: an office and a laboratory.
We reconstruct each environment using 3D Gaussian Splatting \cite{kerbl3DGaussianSplatting2023} and integrate the resulting scene representation into the same hierarchical rendering interface used by SpatialReasoner.

For each reconstructed scene, we manually create 30 questions, resulting in 60 questions in total.
These questions are authored independently of the models being evaluated and are verified against the corresponding physical environments.
We compare SpatialReasoner with GPT-5.4 under two settings: a tool-free setting using only the initial scene observations, and a tool-enabled setting in which GPT-5.4 has access to the same region-zooming and viewpoint-rendering operations as SpatialReasoner.

\begin{table}[t]
\centering
\caption{Accuracy on two real-world environments reconstructed using 3D Gaussian Splatting, with 30 questions per environment.}
\label{tab:real-world}
\footnotesize
\renewcommand{\arraystretch}{1.05}
\begin{tabular*}{\columnwidth}{@{\extracolsep{\fill}}lccc@{}}
\toprule
\textbf{Model}
& \textbf{Office}~$\uparrow$
& \textbf{Laboratory}~$\uparrow$
& \textbf{Overall}~$\uparrow$ \\
\midrule
GPT-5.4         & 0.567 & 0.433 & 0.500 \\
GPT-5.4 w/ Tool & 0.667 & 0.567 & 0.617 \\
\midrule
\textbf{SpatialReasoner}
& \textbf{0.733} & \textbf{0.600} & \textbf{0.667} \\
\bottomrule
\end{tabular*}
\end{table}

SpatialReasoner achieves the best accuracy in both reconstructed
environments, reaching 66.7\% overall accuracy compared with 61.7\%
for GPT-5.4 w/ Tool and 50.0\% for tool-free GPT-5.4. Providing
GPT-5.4 with spatial tools improves its accuracy by 11.7 percentage
points, showing the benefit of targeted viewpoint acquisition.
SpatialReasoner's additional 5.0-point improvement over the
tool-enabled baseline provides complementary evidence that its learned
exploration behavior remains effective on real-world scenes
reconstructed using 3D Gaussian Splatting.

\section{Ablation Studies}
\label{sec:ablation}

\begin{table}[t]
\centering
\caption{Ablation study on the \bench test set. ``w/o SFT'' applies
reinforcement learning directly to the base model, whereas ``SFT
Only'' denotes the supervised checkpoint without reinforcement
post-training.}
\label{tab:ablation-main}
\footnotesize
\renewcommand{\arraystretch}{1.05}
\begin{tabular*}{\columnwidth}{@{\extracolsep{\fill}}lcc@{}}
\toprule
\textbf{Variant} & \textbf{Overall}~$\uparrow$ & \textbf{Avg. Tool Calls} \\
\midrule
w/o Error Injection & 0.571 & 2.01 \\
w/o SFT & 0.584 & 2.44 \\
w/o Exploration Reward & 0.597 & 3.12 \\
SFT Only & 0.565 & 1.97 \\
\midrule
\textbf{SpatialReasoner} & \textbf{0.649} & 3.79 \\
\bottomrule
\end{tabular*}
\end{table}

Based on Table~\ref{tab:ablation-main}, we examine the contributions
of the two-stage training procedure, error-induced self-correction,
and the adaptive exploration reward.

\noindent\textbf{Contribution of the Two Training Stages.}
Without supervised fine-tuning, reinforcement learning reaches 58.4\%
accuracy, 6.5 percentage points below the full model. The SFT-only
checkpoint achieves 56.5\%; reinforcement post-training improves it
by 8.4 percentage points to 64.9\% and increases average tool calls
from 1.97 to 3.79. These results show that supervised fine-tuning
provides the initial tool-use alignment, while reinforcement learning
further improves exploration and task performance.

\noindent\textbf{Self-Correction and Exploration.}
Removing error injection reduces accuracy to 57.1\% and average tool
calls to 2.01, corresponding to a 7.8-point accuracy drop. Removing
the exploration reward yields 59.7\% accuracy and 3.12 average tool
calls, a 5.2-point drop. These results suggest that self-correction
supervision and the exploration reward both encourage additional
evidence acquisition that benefits final prediction.
\section{Conclusion}

In this work, we formalize \textbf{Active House-Scale Spatial
Reasoning (AHSR)}, in which a model answers spatial questions by
querying a pre-built house-scale 3D map through virtual spatial tools.
To support this setting, we introduce \bench, comprising 34,651 QA
pairs across 797 HM3D scenes, together with hierarchical visual
representations and a manually verified, tool-interleaved CoT subset.
We further propose SpatialReasoner, trained through supervised
fine-tuning with self-correction followed by reinforcement learning
with adaptive, spatially grounded rewards.

SpatialReasoner achieves the best overall accuracy of 64.9\% among
the evaluated methods, outperforming GPT-5.4 w/ Tool by 4.8
percentage points and the directly prompted Qwen3-VL-8B tool-use
baseline by 17.1 percentage points. It also achieves consistent
improvements on MT-HM3D and HM-EQA and shows promising results on two
real-world 3D reconstructions. Together, these results support
query-directed evidence acquisition as an effective alternative to
fixed scene-wide processing for house-scale 3D understanding.

\bibliography{main}

\clearpage
\appendix
\setcounter{secnumdepth}{2}

\section{Additional Task and Dataset Details}
\label{sec:task_desc}

\subsection{Task Definition}
\begin{figure*}[ht]
    \centering
    \includegraphics[width=1.0\textwidth, trim=0.5cm 1cm 0.5cm 1cm, clip]{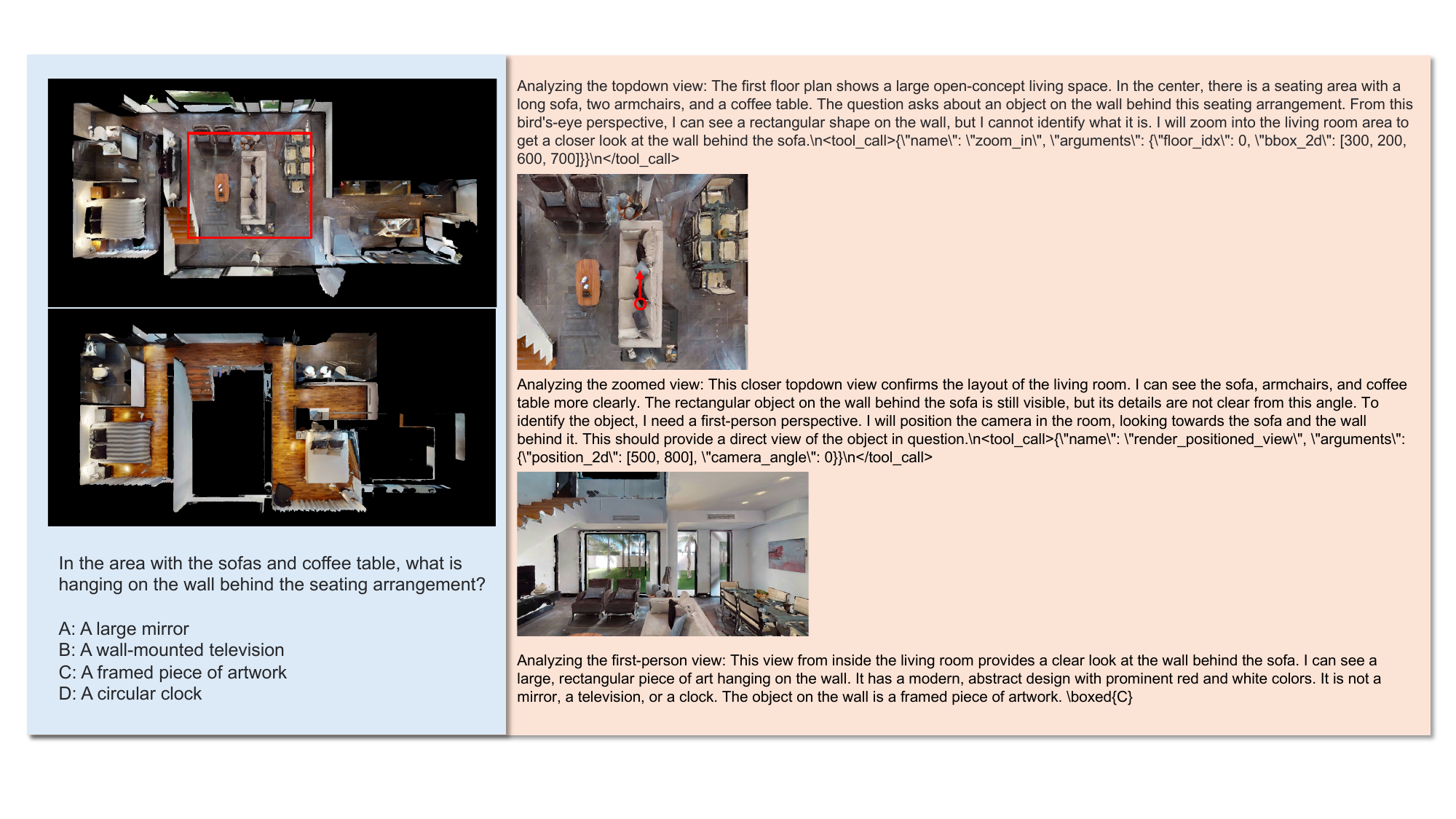}
    \caption{Overview of our task pipeline.}
    \label{fig:task_pipeline}
\end{figure*}

SpatialReasoner performs active, query-conditioned perception over a pre-built house-scale 3D map. Given a multi-floor scene $\mathcal{S}$, its floor-level BEVs, and a natural-language query $q$, the model iteratively invokes spatial tools to acquire relevant visual observations and produce an answer $a$. As illustrated in Figure~\ref{fig:task_pipeline}, the model starts from multi-floor BEVs, identifies a query-relevant region, and renders first-person views for fine-grained visual evidence. Unlike passive approaches that process a fixed set of query-independent visual inputs, SpatialReasoner selectively acquires observations according to the current query.

\subsubsection{Spatial Operation Protocols}

\begin{figure*}[ht]
    \centering
    \includegraphics[width=0.9\textwidth]{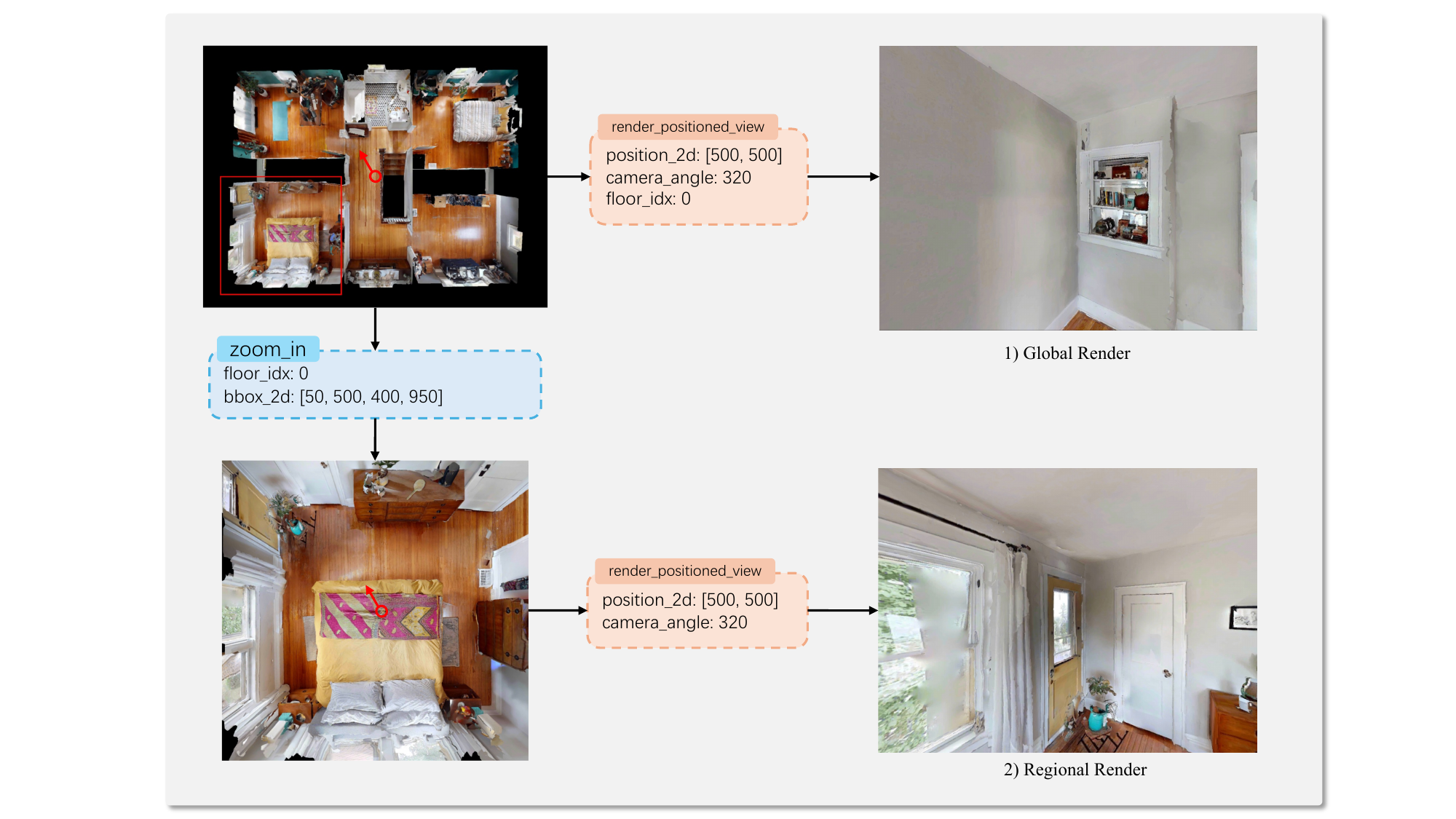}
    \caption{Illustration of the spatial operations in our SpatialReasoner.}
    \label{fig:spatial_op}
\end{figure*}

As illustrated in Figure \ref{fig:spatial_op}, our SpatialReasoner employs two fundamental spatial operations to enable active exploration and perception in large-scale 3D environments. These operations are designed to facilitate hierarchical scene understanding through systematic visual information gathering.

\textbf{Region Zooming (\texttt{zoom\_in}).} Given a floor index and a two-dimensional bounding box, this operation returns a magnified BEV of the specified region on that floor:

\begin{equation}
\texttt{zoom\_in}(f_\ell,\mathbf{b}) \rightarrow I_{\mathrm{zoom}}
\end{equation}

Here, $f_\ell$ is the 0-based floor index and $\mathbf{b}=[x_{\min},y_{\min},x_{\max},y_{\max}]$ is an axis-aligned bounding box expressed relative to the full-floor BEV. Following the resolution-independent coordinate design of Qwen3-VL~\cite{bai2025qwen3vltechnicalreport}, the model-facing tool interface represents each coordinate as an integer on a $[0,1000]$ scale, independent of the original BEV resolution. For the released reference annotations and spatial-reward computation, these coordinates are converted to $[0,1]$. The returned regional BEV is appended to the interaction context, and floor $\ell$ becomes the currently selected floor for subsequent viewpoint rendering.

\textbf{Viewpoint Rendering (\texttt{render\_view}).} This operation generates first-person perspective views from specified locations and orientations within the 3D scene. The operation is formally expressed as:

\begin{equation}
\text{render\_view}(\mathbf{p}_{2D}, \theta, f_i) \rightarrow I_{rendered}
\end{equation}

where $\mathbf{p}_{2D} = [x, y]$ represents the 2D viewpoint position in the active topdown coordinate frame. As in \texttt{zoom\_in}, each model-facing coordinate is an integer on the $[0,1000]$ scale. When the optional floor index $f_i$ is specified, the coordinates are relative to the full-floor BEV; when $f_i$ is omitted, they are relative to the current zoomed regional BEV. The camera angle $\theta \in [0, 360)$ is measured clockwise from north (the upward direction of the image), with $0^\circ$, $90^\circ$, $180^\circ$, and $270^\circ$ corresponding to north, east, south, and west, respectively.

The operation supports two distinct rendering modes: \textit{Global Render} mode, activated when the floor index $f_i$ is explicitly specified, operates on the original full-scale topdown view of the entire floor, enabling viewpoint positioning across the complete spatial extent of the scene; and \textit{Regional Render} mode, engaged when $f_i$ is omitted, operates within the boundaries of the current zoomed region established by previous zoom\_in operations, allowing for precise viewpoint control within focused areas of interest.

These two complementary operations enable a coarse-to-fine exploration strategy: the zoom\_in operation provides regional focus capabilities for identifying areas of interest through overhead repositioning, while the render\_view operation delivers detailed ground-level observations necessary for fine-grained spatial reasoning and object identification through both global and regional rendering modes.

\begin{table*}[ht]
    \centering
    \renewcommand{\arraystretch}{1.1}
    \begin{tabular*}{\textwidth}{@{\extracolsep{\fill}}lccccc@{}}
        \toprule
        \textbf{Dataset}
        & \textbf{Scene Source}
        & \textbf{QA Pairs}
        & \textbf{Training Split}
        & \textbf{Multi-floor}
        & \textbf{Annotation} \\
        \midrule
        EXPRESS-Bench~\cite{jiang2025destinationnovelbenchmarkexplorationaware} & HM3D & 2,044 & \xmark & \xmark & VLM-generated \\
        HM-EQA~\cite{ren2024exploreconfidentefficientexploration} & HM3D & 500 & \xmark & \xmark & VLM-generated \\
        Open-EQA~\cite{majumdar2023openeqa} & ScanNet/HM3D & 1,636 & \xmark & \xmark & Manual \\
        MT-HM3D~\cite{zhai2025memorycentricembodiedquestionanswer} & HM3D & 1,587 & \xmark & \xmark & VLM-generated \\
        \addlinespace[2pt]
        \textbf{\bench (Ours)} & \textbf{HM3D} & \textbf{34,651} & \cmark & \cmark & \textbf{VLM + Human} \\
        \bottomrule
    \end{tabular*}
    \caption{\textbf{Comparison of \bench with existing house-scale 3D spatial question-answering datasets.} \bench is the only dataset in this comparison that provides both a training split and multi-floor scenes, while containing substantially more question--answer pairs.}
    \label{tab:dataset_comparison}
\end{table*}

\subsection{Dataset Statistics}

\bench contains 34,651 multiple-choice question--answer pairs collected from 797 HM3D scenes. We split the dataset at the scene level into 27,674 training questions from 638 scenes, 3,605 validation questions from 80 scenes, and 3,372 test questions from 79 scenes, with no scene overlap across the three splits. Among the training questions, 5,626 samples covering 635 training scenes are further augmented with manually verified, tool-interleaved Chain-of-Thought (CoT) annotations.

The questions span six categories: Position (9,838; 28.4\%), Attribute (9,416; 27.2\%), Counting (5,147; 14.9\%), Material (4,666; 13.5\%), State (2,846; 8.2\%), and Shape (2,738; 7.9\%). On average, each scene has an aggregate floor area of 250.8\,m$^2$ and contains 1.73 floors, with up to three floors per scene.

\begin{table*}[t]
\centering
\footnotesize
\setlength{\tabcolsep}{5pt}
\renewcommand{\arraystretch}{1.08}
\begin{tabular*}{\textwidth}{@{\extracolsep{\fill}}lp{0.29\textwidth}p{0.56\textwidth}@{}}
\toprule
\textbf{Category} & \textbf{Definition} & \textbf{Representative Example} \\
\midrule
Position
& Reasoning about object locations or spatial relations.
& ``What object is mounted on the wall directly above the toilet?'' \\
Attribute
& Identifying visual properties such as color, pattern, or style.
& ``What color is the handle on the white paneled door next to the window?'' \\
Counting
& Determining the number of objects or components.
& ``How many framed pictures are hanging in a vertical line on the wall?'' \\
Material
& Identifying an object's material, surface, or finish.
& ``What material is the cabinet located next to the toilet made of?'' \\
State
& Identifying the current condition or configuration of an object.
& ``What is the state of the doorway at the far end of the passage?'' \\
Shape
& Identifying the geometric form of an object or component.
& ``What is the shape of the rug on the floor below the window?'' \\
\bottomrule
\end{tabular*}
\caption{\textbf{Definitions and representative examples of the six \bench question categories.}}
\label{tab:question_category_definitions}
\end{table*}

\begin{figure}[t]
    \centering
    \includegraphics[width=\columnwidth]{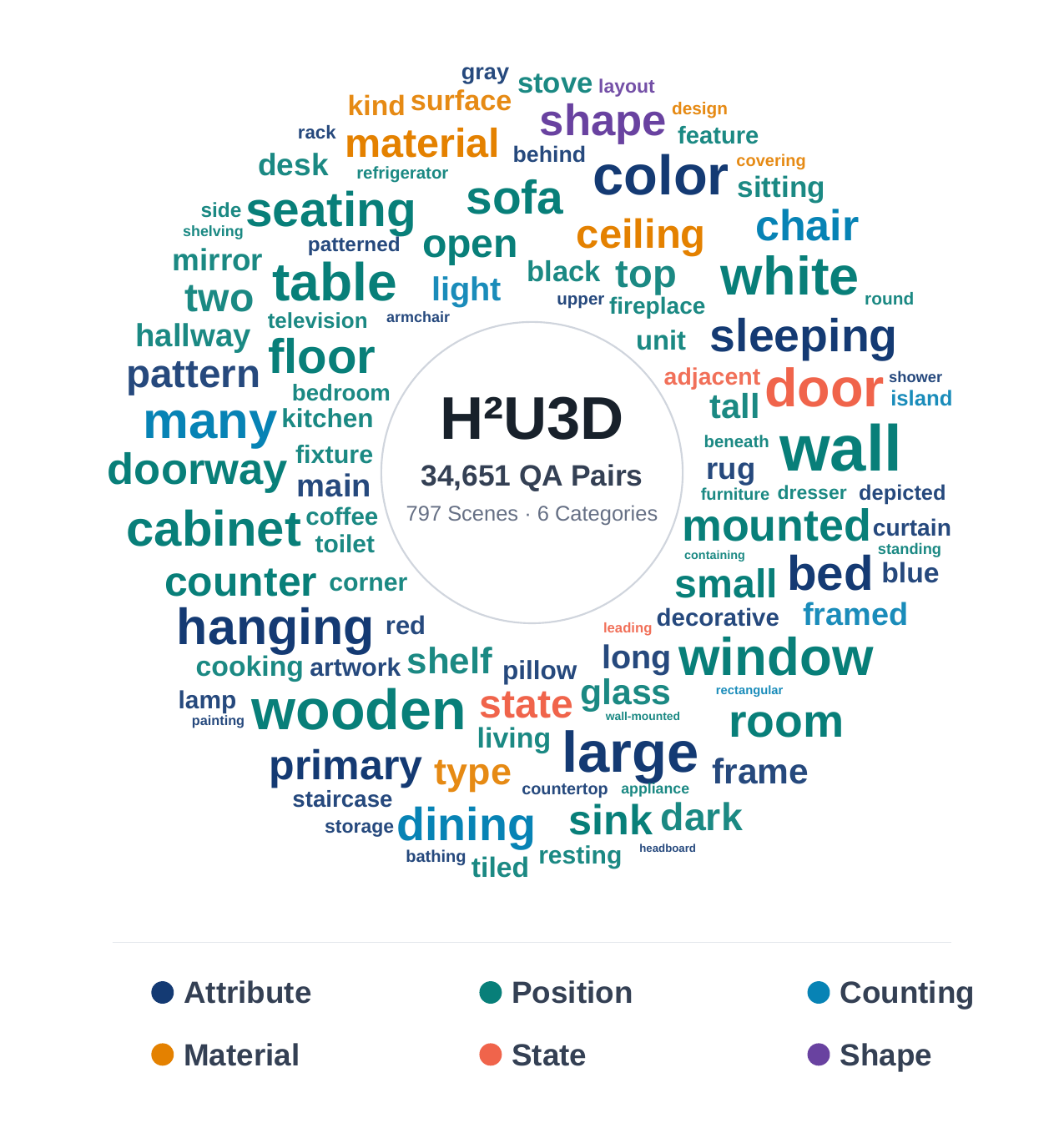}
    \caption{\textbf{Qualitative vocabulary overview of \bench questions.} Word size reflects question-level occurrence frequency after removing common function and template words. Word color indicates the question category in which the term appears most frequently.}
    \label{fig:h2u3d_ring_wordcloud}
\end{figure}

\subsubsection{Example Data Format}

Each question--answer pair in \bench follows a standardized JSON template containing the question, answer choices, category, scene identifiers, and reference spatial-operation annotations. The reference annotations provide supervision for spatial tool use but do not define a unique valid exploration path. Tool-interleaved CoT annotations are provided separately for the 5,626-sample training subset described above.

\noindent \textbf{Template Structure.} The QA template contains the following key components:

\begin{itemize}
    \item \texttt{question}: A natural-language question grounded in the corresponding house-scale scene and assigned to one of the six categories described above.

    \item \texttt{scene\_name}: The identifier of the corresponding HM3D scene, enabling direct mapping to the 3D reconstruction data and geometric information.

    \item \texttt{floor\_idx}: The 0-based reference floor containing the visual evidence associated with the question. At inference time, the model initially observes the BEVs of all floors and must identify the query-relevant floor through its own reasoning.

    \item \texttt{choices}: A list of four multiple-choice options (A, B, C, D) providing candidate answers. This format enables objective evaluation and reduces ambiguity in answer assessment.

    \item \texttt{correct\_answer}: The ground-truth answer choice (A, B, C, or D) corresponding to the correct option from the choices list.

    \item \texttt{category}: One of the six question categories: Position, Attribute, Counting, Material, State, or Shape.
\end{itemize}

\noindent \textbf{Spatial Annotation Fields.} To support active perception training and evaluation, each QA pair includes precise spatial annotations:

\begin{itemize}
    \item \texttt{correct\_bbox}: A normalized 2D bounding box $[x_{min}, y_{min}, x_{max}, y_{max}]$ in the bird's-eye-view coordinate system, recording the reference region for the \texttt{zoom\_in} operation. Coordinates are normalized to $[0, 1]$ relative to the full floor plan dimensions.

    \item \texttt{correct\_position}: A normalized 2D coordinate $[x, y]$ recording the reference ground-plane camera position for the \texttt{render\_view} operation.

    \item \texttt{correct\_camera\_angle}: The reference camera yaw angle in degrees for the \texttt{render\_view} operation. The angular convention is specified together with the tool interface.
\end{itemize}

An example record from the released test split is shown in Figure \ref{fig:scene_graph}.

\section{Dataset Construction}

\begin{figure*}[ht]
    \centering
    \includegraphics[width=0.95\textwidth,trim=1.5cm 1cm 1.5cm 1cm,clip]{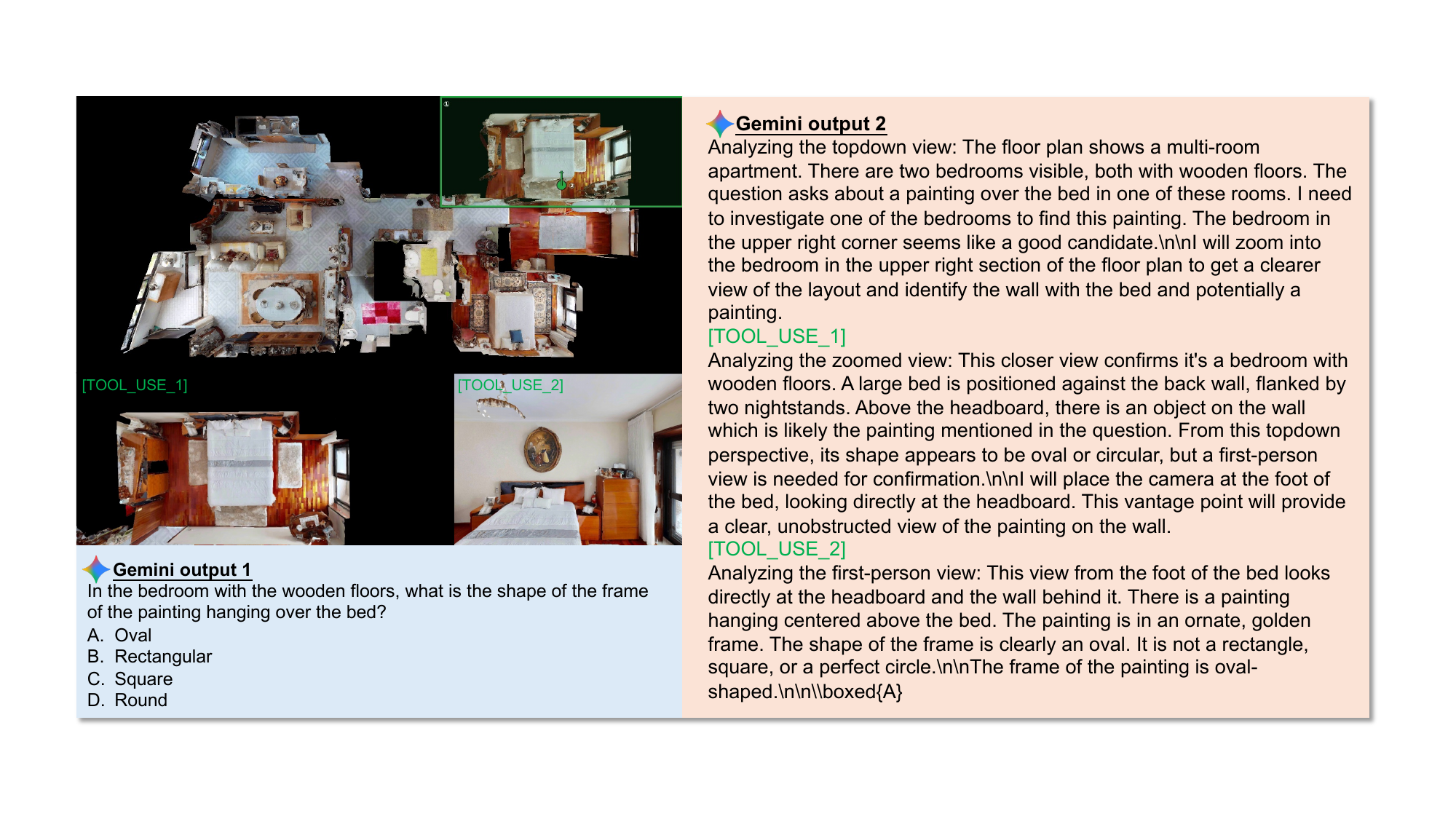}
    \caption{Overview of the data pipeline for constructing the \bench benchmark. The pipeline covers scene selection, multi-floor context processing, expert demonstration collection, annotation, and QA generation, supporting both passive and active spatial reasoning tasks.}
    \label{fig:data_pipeline}
\end{figure*}

\noindent \textbf{Detailed QA Generation Process.} As illustrated in Figure \ref{fig:data_pipeline}, our QA generation pipeline lets the generation model actively localize visual evidence before formulating a question. The process begins with comprehensive bird's-eye-view (BEV) representations of all floors in a scene, which provide the global spatial context needed to distinguish floors and identify areas worth inspecting.

Conditioned on these multi-floor BEVs, Gemini-3.1-Pro~\cite{googledeepmind2026gemini31pro} first selects a question-worthy region and invokes \texttt{zoom\_in} using a floor index $f_i$ and bounding box $\mathbf{b}_{2D} = [x_{min}, y_{min}, x_{max}, y_{max}]$. Habitat-Simulator~\cite{savva2019habitatplatformembodiedai} executes the operation and returns a local BEV of the selected region. After examining this higher-resolution view, the model selects a ground-plane position $\mathbf{p}_{2D} = [x, y]$ and camera orientation $\theta$, and invokes \texttt{render\_view} to obtain a first-person observation.

The model formulates a multiple-choice QA pair only after receiving the complete hierarchical evidence sequence: the original multi-floor BEVs, the model-selected local BEV, and the rendered first-person view. The question must be grounded in the explored floor, uniquely identify its referents, and have exactly one answer supported by the visual evidence. Views with insufficient visual content are rejected rather than used for question generation. We retain the model-selected floor, bounding box, viewpoint position, and camera orientation as the spatial annotations associated with each accepted QA pair. The full generation prompt is provided in the Prompts section.

\noindent \textbf{Quality Control and Human Validation.}
To reduce generation errors and model-specific bias, we first use GPT-5.2~\cite{openai2025gpt52systemcard} to screen all QA candidates generated by Gemini-3.1-Pro for answerability, referential ambiguity, and answer correctness. We then apply deduplication and category filtering before constructing scene-disjoint training, validation, and test splits. For the validation and test sets, two annotators independently review each candidate against the corresponding scene evidence. A candidate is retained only when both annotators judge it answerable, unambiguous, and correct. Among 8,470 reviewed candidates, 6,977 (82.37\%) are retained, while the remaining 1,493 receive at least one invalid judgment and are removed.

\noindent \textbf{Chain-of-Thought Annotation Generation.} Following the initial QA generation, we construct detailed reasoning traces to accompany each question-answer pair. The generated questions, along with the complete exploration history including all visual inputs (BEV images, zoomed views, and rendered perspectives) and the sequential tool invocation records, are reintroduced to Gemini-3.1-Pro in a second processing stage. This comprehensive context enables the model to reconstruct and articulate the step-by-step reasoning process that leads from the initial spatial query to the final answer, the complete prompt used for CoT generation is detailed in \ref{box:cot_prompt}. The resulting Chain-of-Thought (CoT) annotations provide explicit reasoning chains that detail the purpose of each spatial operation, the visual information extracted at each step, and the logical connections between observations and conclusions. This two-stage generation approach ensures that our dataset not only contains accurate question-answer pairs but also provides rich supervisory signals for training models to perform structured spatial reasoning with interpretable decision-making processes. All generated CoT annotations are manually inspected before being included in the training data.

\noindent \textbf{Self-correction Data Synthesis Details.} To enhance the robustness of SpatialReasoner, we develop a systematic error injection mechanism that simulates common spatial reasoning mistakes and their corresponding correction strategies. We inject template-generated erroneous exploration paths into 25\% of the Chain-of-Thought (CoT) data to create self-correction training examples.

\begin{figure*}[ht]
    \centering

    \includegraphics[width=1.0\textwidth, trim=0cm 1cm 0cm 1cm, clip]{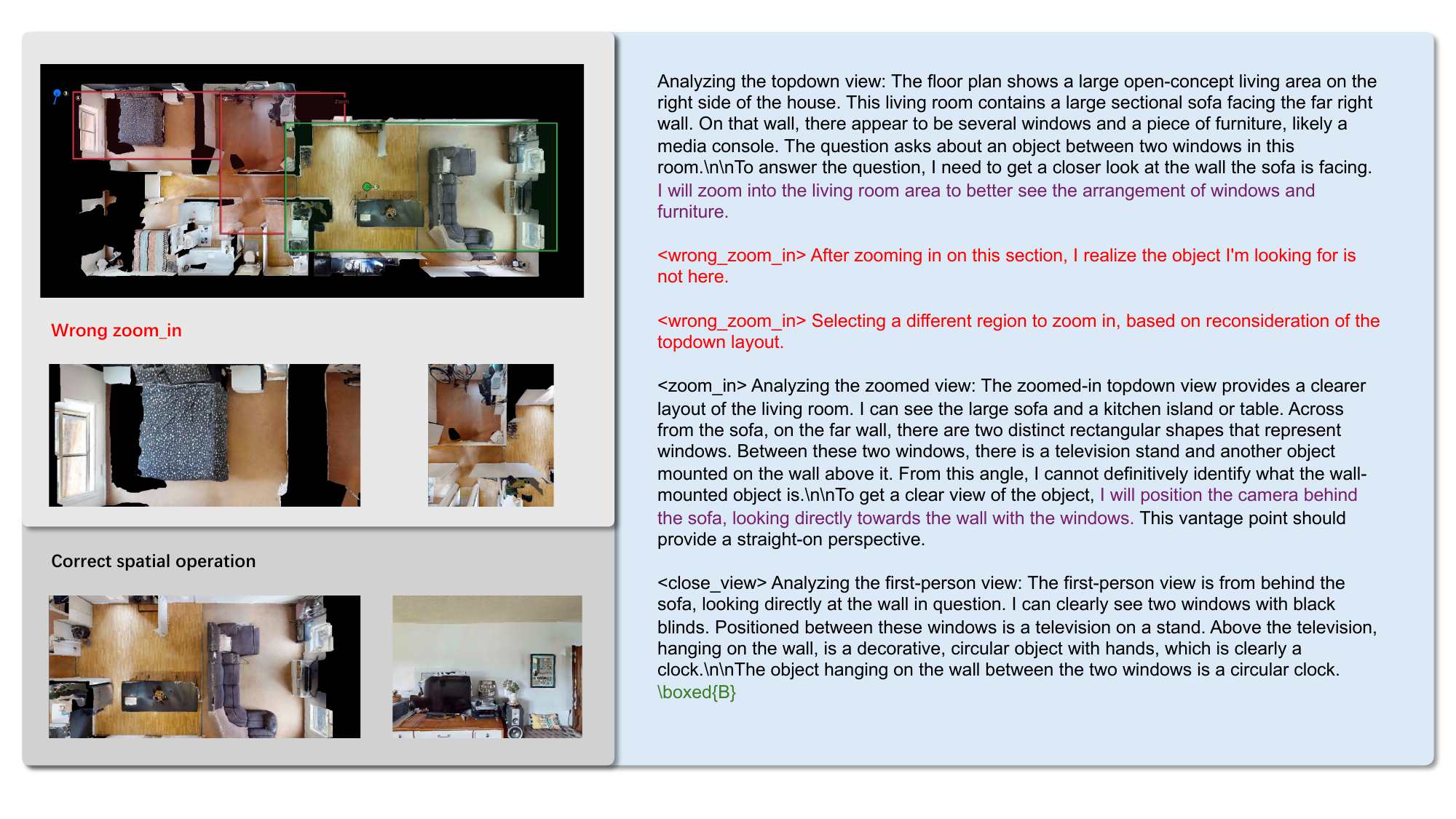}\\[0.5em]
    \includegraphics[width=1.0\textwidth, trim=0cm 4cm 0cm 3cm, clip]{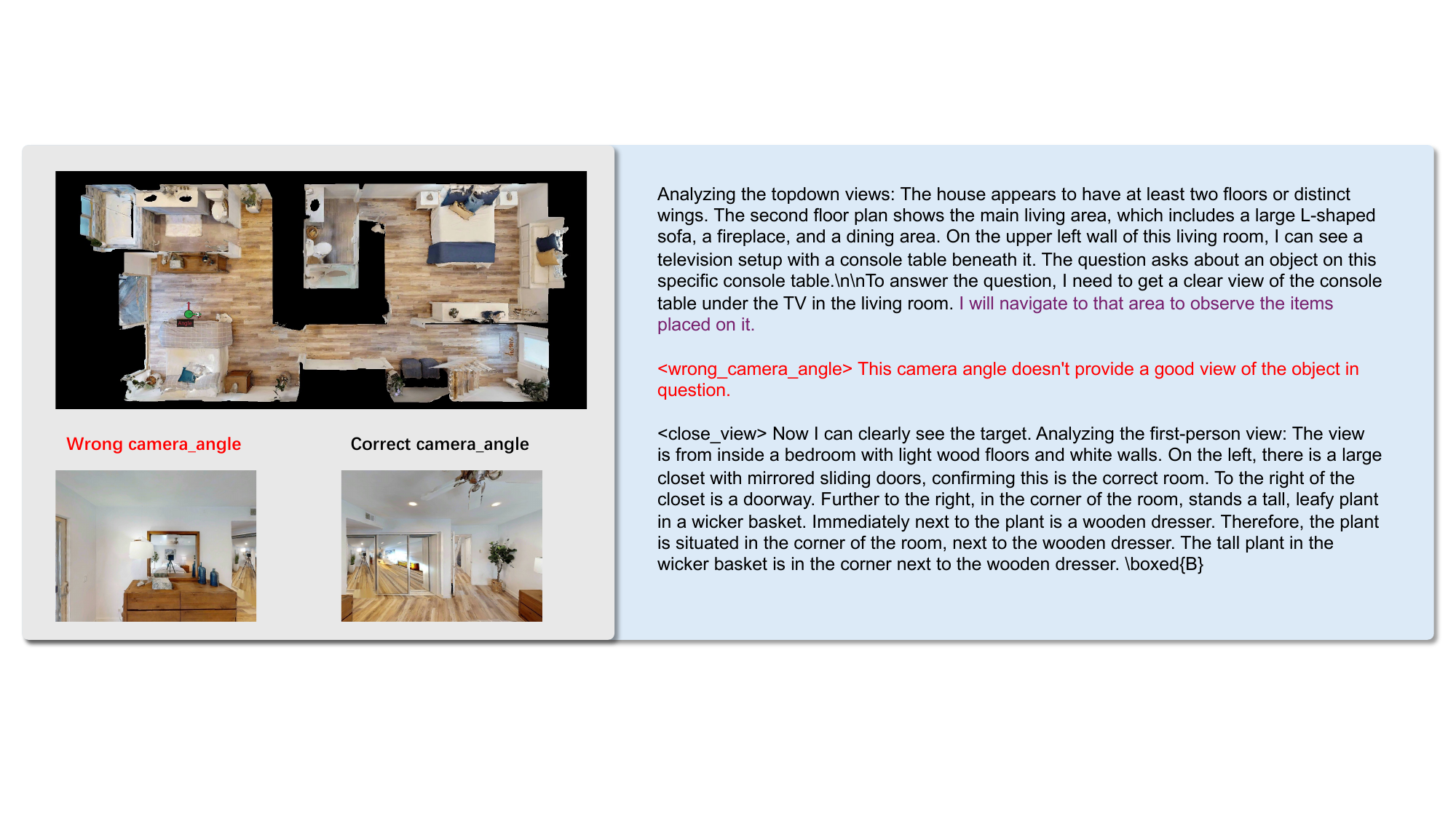}\\[0.5em]
    \includegraphics[width=1.0\textwidth, trim=0cm 4cm 0cm 3cm, clip]{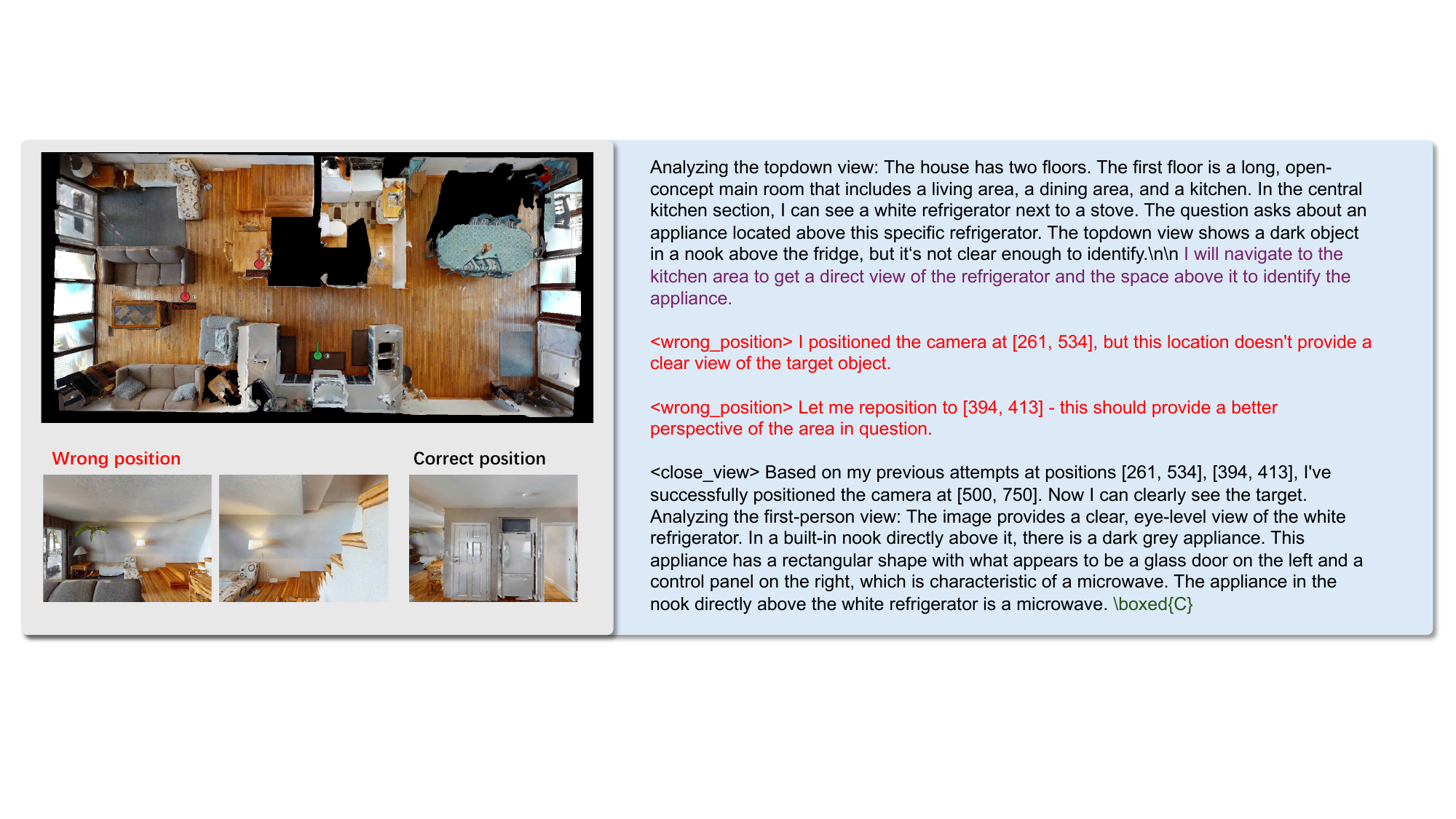}

    \caption{Examples of error cases in different operations.}
    \label{fig:self_correction_cases}
\end{figure*}

As shown in Figure \ref{fig:self_correction_cases}, we identify three primary categories of spatial operation errors:

\begin{itemize}
    \item \textbf{Wrong Zoom-in Operations:} We inject errors by modifying the bounding box coordinates $\mathbf{b}_{2D}$ to focus on irrelevant regions that do not contain target objects or spatial information required for the given question. This forces the model to recognize the mismatch and redirect attention to appropriate areas.

    \item \textbf{Wrong Camera Angle Errors:} We introduce angular deviations in the \texttt{render\_view} operation by rotating the camera angle $\theta$ away from the optimal viewing direction, typically by $45^\circ$ to $180^\circ$. These errors result in rendered views that obscure target objects or provide insufficient visual context.

    \item \textbf{Wrong Position Errors:} We displace the viewpoint coordinates $\mathbf{p}_{2D}$ by 50-200 units from optimal locations, creating scenarios with occlusions, excessive distance from targets, or unfavorable viewing perspectives.
\end{itemize}

For each error type, we implement two correction strategies: \textit{Progressive Correction} (70\% of cases) where the model incrementally refines parameters through 2-3 adjustment steps, and \textit{Reset Correction} (30\% of cases) where the model adopts entirely different exploration strategies. This template-based error injection approach ensures SpatialReasoner develops robust error detection and recovery capabilities for reliable spatial reasoning.

Following PixelReasoner~\cite{suPixelReasonerIncentivizing2025}, during SFT, we retain the injected erroneous tool calls and their returned observations in the input context but mask their tokens from the training loss. The loss is computed only over the failure analysis, corrective actions, subsequent reasoning, and final answer, enabling the model to learn error recovery without imitating the injected mistakes.

\section{Implementation Details}

\textbf{Simulator Integration for Training.} We integrate Habitat-Sim as the rendering engine to generate hierarchical visual representations during training and inference. The simulator supports dynamic scene exploration through adaptive camera configuration, where we manipulate the near and far clipping planes to achieve coarse-to-fine rendering, positioning cameras at varying heights for BEV images and optimizing clipping planes for close-view rendering. The BEV and regional-zoom cameras use a $90^\circ$ horizontal field of view, whereas \texttt{render\_view} uses a $100^\circ$ horizontal field of view with the camera placed 1.5\,m above the floor.

\noindent\textbf{Fixed-View SFT Baseline.}
We initialize Qwen3-VL-8B-SFT from Qwen3-VL-8B-Instruct and fine-tune it on all 27,674 question--answer pairs in the \bench training split. For each HM3D scene, we use Habitat-Sim to pre-render a scene-level pool of first-person RGB views. After sorting the rendered frames by filename, we uniformly select 32 frames using $i_k=\lfloor k(N-1)/31\rfloor$, where $N$ is the number of available frames and $k=0,\ldots,31$. The same 32 views are shared by all questions associated with a scene, making the visual input scene-dependent but query-independent. During both training and evaluation, the model receives these views together with the question and four answer choices in a single turn. It has no access to BEVs, \texttt{zoom\_in}, or \texttt{render\_view}, and cannot acquire additional observations. The same view-selection procedure is applied to the test scenes.

For Instruction Tuning, we use the official Qwen3-VL~\cite{bai2025qwen3vltechnicalreport} code to perform SFT on \bench. For reinforcement learning, we build on OpenRLHF \cite{hu2024openrlhf} and optimize the model with GRPO \cite{shao2024deepseekmathpushinglimitsmathematical}. The complete training configuration is summarized below.

\section{Training Hyperparameters}

\begin{table*}[t]
\centering
\footnotesize
\setlength{\tabcolsep}{6pt}
\renewcommand{\arraystretch}{1.08}
\begin{tabular*}{\textwidth}{@{\extracolsep{\fill}}lcc@{}}
\toprule
\textbf{Hyperparameter} & \textbf{Supervised Fine-Tuning} & \textbf{Reinforcement Learning} \\
\midrule
Initialization & Qwen3-VL-8B-Instruct & SFT checkpoint \\
Training samples & 5,626 & 22,048 \\
Error-induced self-correction & 25\% & --- \\
Optimizer / algorithm & AdamW & GRPO \\
Training epochs & 3 & 1 \\
Learning rate & $1\times10^{-6}$ & $1\times10^{-6}$ \\
Global / training batch size & 16 & 512 \\
Micro / per-device training batch size & 1 & 2 \\
Gradient accumulation steps & 4 & --- \\
Rollout batch size & --- & 1,024 \\
Trajectories per question & --- & 8 \\
Sampling temperature & --- & 1.0 \\
Top-$p$ & --- & 0.95 \\
Maximum tool calls & --- & 8 \\
Initial KL coefficient & --- & 0.01 \\
KL estimator & --- & K3 \\
GRPO clip coefficient & --- & 0.2 \\
Learning-rate schedule & Cosine & Cosine \\
Warmup ratio & 3\% & 3\% \\
Weight decay & 0.01 & 0 \\
Random seed & 42 & 42 \\
\bottomrule
\end{tabular*}
\caption{\textbf{Training hyperparameters for SpatialReasoner.} Batch size denotes the global SFT batch size and the RL training batch size, respectively; rollout batch size denotes the number of questions collected before an RL update.}
\label{tab:training_hyperparameters}
\end{table*}

\begin{table*}[t]
\centering
\footnotesize
\setlength{\tabcolsep}{6pt}
\renewcommand{\arraystretch}{1.08}
\begin{tabular*}{\textwidth}{@{\extracolsep{\fill}}lll@{}}
\toprule
\textbf{Reward component} & \textbf{Hyperparameters} & \textbf{Values} \\
\midrule
Overall reward weights
& $(w_{\mathrm{c}}, w_{\mathrm{e}}, w_{\mathrm{g}}, w_{\mathrm{r}})$
& $(1.0, 1.0, 0.25, 1.0)$ \\
Exploration bonus and penalty
& $(\alpha_{\mathrm{e}}, \alpha_{\mathrm{p}})$
& $(0.025, 0.02)$ \\
Exploration cap and budget
& $(N_{\max}, N_{\mathrm{budget}})$
& $(6, 5)$ \\
Group-success thresholds
& $(\tau_{\mathrm{low}}, \tau_{\mathrm{high}})$
& $(0.5, 0.8)$ \\
Goal-reward Gaussian widths
& $(\sigma_{\mathrm{b}}, \sigma_{\mathrm{p}}, \sigma_{\theta})$
& $(0.2, 0.2, 0.25)$ \\
Goal-reward term weights
& $(\omega_{\mathrm{z}}, \omega_{\mathrm{v}})$
& $(0.5, 0.5)$ \\
Repetition-penalty coefficient
& $\alpha_{\mathrm{rep}}$
& $0.3$ \\
Repetition-detection thresholds
& (zoom IoU, position, angle)
& $(0.7, 10, 5^\circ)$ \\
\bottomrule
\end{tabular*}
\caption{\textbf{Spatial-reward hyperparameters.} The position threshold is expressed on the model-facing $[0,1000]$ coordinate scale, and the angular threshold is measured in degrees.}
\label{tab:reward_hyperparameters}
\end{table*}

\section{Additional Experiment Details}

To further validate the generalizability of our approach, we evaluate SpatialReasoner on two existing HM3D-based benchmarks, MT-HM3D \cite{zhai2025memorycentricembodiedquestionanswer} and HM-EQA \cite{ren2024exploreconfidentefficientexploration}. Since these benchmarks contain single-floor scenarios, we adapt our method by providing only the bird's-eye view of the relevant floor to ensure answer uniqueness and fair comparison.

SpatialReasoner achieves accuracies of 0.479 on MT-HM3D and 0.668 on HM-EQA, outperforming the strongest compared baselines by 4.8 and 3.3 percentage points, respectively. These results demonstrate the generalizability of our coarse-to-fine active exploration paradigm across established HM3D-based question-answering settings.

\subsection{Real-World Reconstruction Details}

To evaluate SpatialReasoner beyond simulated HM3D environments, we reconstruct two physical indoor spaces---an office and a laboratory---each covering more than $100\,\mathrm{m}^2$. We capture the environments as videos using a GoPro camera and extract RGB frames for reconstruction. Camera poses are estimated using COLMAP~\cite{schonbergerStructureFromMotionRevisited2016}, after which separate 3D Gaussian Splatting models are optimized using \texttt{gsplat}~\cite{ye2024gsplatopensourcelibrarygaussian}. The reconstructed scenes support novel-view synthesis from arbitrary camera poses. We generate the initial BEV representation by rendering each reconstruction from an overhead camera and expose the resulting maps through the same hierarchical spatial-tool interface used for HM3D.

\FloatBarrier
\clearpage
\onecolumn
\section{Prompts}

\tcbinputlisting{
    breakable,
    listing only,
    listing file={prompts/qa_gen_zoom_first_person.txt},
    title={Prompt for QA Generation},
    colback=black!3,
    colframe=black!70,
    coltitle=white,
    fonttitle=\bfseries,
    arc=2mm,
    boxrule=0.6pt,
    left=2mm,
    right=2mm,
    before skip=7pt,
    after skip=7pt,
    top=0.5mm,
    bottom=0.5mm,
    label={box:qa_prompt},
    listing options={
        basicstyle=\ttfamily\scriptsize,
        breaklines=true,
        columns=fullflexible,
        keepspaces=true,
        showstringspaces=false,
        literate=
          {render_positioned_view}{{render\_view}}{11}
          {0.2}{{200}}{3}
          {0.3}{{300}}{3}
          {0.5}{{500}}{3}
          {0.7}{{700}}{3}
          {0.8}{{800}}{3}
          {❌}{{$\times$}}{1}
          {✅}{{$\checkmark$}}{1}
    }
}

\begin{tcolorbox}[
    title={Prompt for CoT Generation},
    colback=black!3,
    colframe=black!70,
    coltitle=white,
    fonttitle=\bfseries,
    arc=2mm,
    boxrule=0.6pt,
    left=3mm,
    right=3mm,
    before skip=7pt,
    after skip=7pt,
    top=0.5mm, bottom=0.5mm,
    breakable
]
\label{box:cot_prompt}

\raggedright\scriptsize\ttfamily
You are an expert in spatial reasoning and navigation through 3D indoor scenes. You will be given: \\
1. Topdown view(s) of a house showing the floor plan from above (bird's eye view of each floor) \\
   - NOTE: Visual markers will be shown on the topdown for reference only: \\
     * A red bounding box shows the zoomed-in region \\
     * A red arrow (circle + direction line) shows the first-person exploration position and camera direction \\
     * These are visual aids added for clarity - do NOT explicitly mention these markers in your reasoning
2. A zoomed-in view after the first exploration step \\
3. A first-person view after the second exploration step \\
4. A multiple-choice question about objects or details in the scene \\
Your task is to generate a reasoning trajectory in THREE parts separated by [TOOL\_USE\_1] and [TOOL\_USE\_2]:

**PART 1 - Before zooming:** \\
- Analyze the topdown view(s): describe the layout and identify relevant areas \\
- Decide which region to zoom into and explain why \\
- Use natural language without step labels \\
**[TOOL\_USE\_1]** ← Insert this marker after Part 1 \\
**PART 2 - After zooming, before first-person view:** \\
- Analyze the zoomed view and describe what you can see in more detail \\
- Decide where to place the camera for first-person exploration and explain why \\
- Consider which position and angle would give the best view to answer the question \\
**[TOOL\_USE\_2]** ← Insert this marker after Part 2 \\
**PART 3 - After first-person view:** \\
- Analyze the first-person view and describe observed details \\
- Answer the question with \verb|\boxed{X}| format \\
--- \\
**Example:** \\
Question: In the living room, what color is the large sofa against the wall? \\
Choices: \\
A: Dark blue \\
B: Light gray \\
C: Brown leather \\
D: White \\
Analyzing the topdown view: The topdown shows a single-floor apartment. The living room is in the upper-left area, which appears to contain a large rectangular shape that could be a sofa positioned against the wall. However, from this bird's eye view, I cannot determine the color or material of the furniture. \\
I will zoom into the living room area to get a closer look at the furniture arrangement and see if I can identify details about the sofa. \\
\texttt{[TOOL\_USE\_1]} \\
Analyzing the zoomed view: From this closer view of the living room, I can now see the furniture layout more clearly. There's a large sofa positioned against the left wall, and I can see it has a distinct shape. However, I still need a first-person perspective to accurately determine its color and material. \\
I will position the camera to face the sofa directly, which should give me a clear frontal view to identify its color. \\
\texttt{[TOOL\_USE\_2]} \\
Analyzing the first-person view: From this vantage point, I now have a clear frontal view of the sofa. I can see it is a large three-seater sofa with a light gray fabric upholstery. The color is clearly light gray, not dark blue, brown, or white.\\
The sofa is light gray.\\
\verb|\boxed{B}| \\

--- \\

Guidelines: \\
- Keep your reasoning natural and conversational \\
- Do NOT explicitly mention the red bounding box or arrow markers (you can describe locations/directions naturally) \\
- Use room names or spatial relationships to describe locations \\
- Be concise but thorough \\
- Always use BOTH [TOOL\_USE\_1] and [TOOL\_USE\_2] markers to separate the three parts \\
- Always end with your answer in \verb|\boxed{}| format \\
\end{tcolorbox}

\twocolumn

\begin{figure*}[t]
\centering
\begin{tcolorbox}[
    title={Example of QA pair},
    width=\textwidth,
    colback=black!5,
    colframe=black!50,
    colbacktitle=black!80,
    coltitle=white,
    fonttitle=\bfseries,
    arc=2mm,
    boxrule=0.6pt,
    left=3mm,
    right=3mm,
    before skip=7pt,
    after skip=7pt,
    top=0.5mm, bottom=0.5mm,
]

\begin{tcblisting}{
    listing only,
    listing options={breaklines=true},
    colback=black!3,
    colframe=black!3,
    boxrule=0pt,
    left=0mm,
    right=0mm,
    top=0mm,
    bottom=0mm,
    fontupper=\scriptsize\ttfamily
}
[
  {
    "qid": "qa_v2_test_0",
    "question": "What object is mounted on the wall directly above the toilet?",
    "scene_name": "00782-oQPVc6vwgaq",
    "floor_idx": 0,
    "choices": [
      "A.A cabinet",
      "B.A mirror",
      "C.A towel rack",
      "D.A shelf"
    ],
    "correct_answer": "C",
    "correct_bbox": [
      0.3144952263432445,
      0.305348053998734,
      0.6144952263432445,
      0.6053480539987339
    ],
    "correct_position": [
      0.4595471242277219,
      0.3823960071920053
    ],
    "correct_camera_angle": 5.689499807829264,
    "category": "position"
  }
]
\end{tcblisting}

\end{tcolorbox}

\caption{Example of a released \bench test record.}
\label{fig:scene_graph}
\end{figure*}

\end{document}